\documentclass[runningheads]{llncs}

\usepackage{graphicx}
\usepackage{float}
\usepackage{multirow}
\usepackage[hyphens]{url}
\usepackage{hyperref}
\hypersetup{colorlinks=false,breaklinks=true}
\pdfoutput=1
\begin{document}
\emergencystretch 3em
\title{Fused Text Recogniser and Deep Embeddings Improve Word Recognition and Retrieval}
\titlerunning{Improved Word Recognition and Retrieval}
\author{Siddhant Bansal \and
Praveen Krishnan \and
C.V. Jawahar}
\authorrunning{Siddhant et al.} 
\institute{Center for Visual Information Technology, IIIT Hyderabad, India
\email{siddhant.bansal@students.iiit.ac.in,}
\email{praveen.krishnan@research.iiit.ac.in, jawahar@iiit.ac.in}}

\maketitle
\begin{abstract}
Recognition and retrieval of textual content from the large document collections
have been a powerful use case for the document image analysis community.
Often the word is the basic unit for recognition as well as retrieval.
Systems that  rely only on the text recogniser’s ({\sc ocr}) output are 
not robust enough in many situations, especially when the word recognition 
rates are poor, as in the case of historic documents or digital libraries. An alternative has been 
word spotting based methods that retrieve/match words based on a 
holistic representation of the word.
In this paper, we fuse the noisy output of text recogniser with a
deep embeddings representation derived out of the entire word.
We use average and max fusion for improving the ranked results in
the case of retrieval.
We validate our methods on a collection of Hindi documents. We 
improve word recognition rate by $1.4\%$ and retrieval by $11.13\%$
in the mAP.




\end{abstract}

\section{Introduction}

Presence of large document collections like the Project Gutenberg \cite{guten} and the Digital Library of India ({\sc dli}) \cite{dli} has provided access to many books in English and Indian languages. 
Such document collections cover a broad range of disciplines like history, languages, art and science, thereby, providing free access to a vast amount of information. 
For the creation of such libraries, the books are converted to machine-readable text by using Optical Character Recognition ({\sc ocr}) solutions.

\begin{figure}
    \centering
    \includegraphics[scale=0.51]{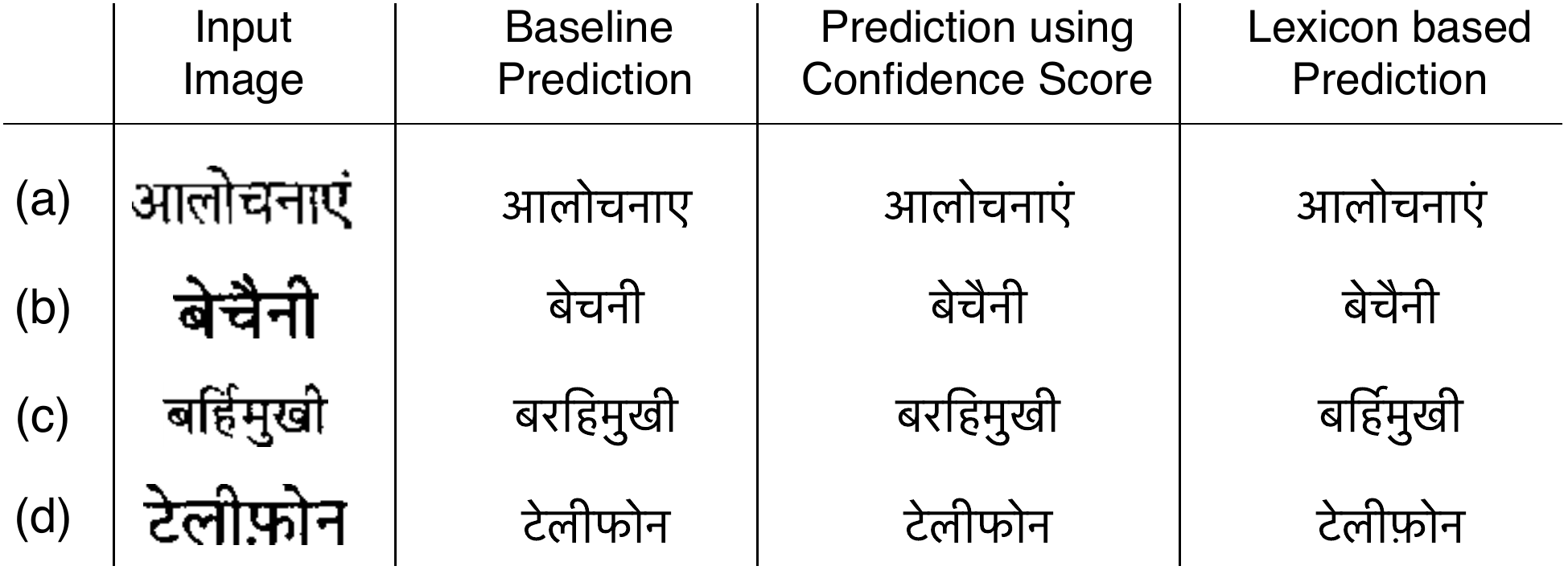}
    \caption{In this figure we show the word recognition results. In (a) and (b), the correct word is selected using both the confidence score and lexicon based methods proposed in this work. In (c) and (d), the correct output is predicted only by the lexicon based prediction. Both the methods are successful in capturing very low level details that are missed by the baseline text recognition methods.}
    \label{fig:retrieved}
\end{figure}{}

Use of {\sc ocr} has also enabled successful retrieval of relevant content in document collections.
In this work, we aim at improving the word recognition and retrieval performance for the Hindi language. 
This is challenging since the data in these libraries consists of
degraded images and complex printing styles.
Current methods providing content-level access to a large corpus can be divided into two classes: (a) text recognition ({\sc ocr}) and (b) word spotting. 
Text recognition-based approaches
have shown to perform well in many situations~\cite{Shi,Pham,Chen}. Using this technique, scanned documents are converted to machine-readable text and then the
search is carried out for the queries on the generated text. 
On the other hand, word spotting is a recognition-free approach for text recognition.
Here, embeddings for the words in the documents are extracted and the nearest neighbour search is performed to get a ranked list.
Various attempts \cite{Bhardwaj,Chaudhury,Meshesha,Shekhar} have been made for creating systems using recognition-free approaches also. 
However, all of these methods fall short in taking advantage of both, i.e., effectively fusing recognition-based and recognition-free methods.

We aim at exploring the complementary behavior of recognition-based and recognition-free approaches.
For that, we generate the predictions by a {\sc crnn}~\cite{Shi} style text recognition network, and get the deep embeddings using the End2End network~\cite{e2e}.
Figure \ref{fig:retrieved} shows the qualitative results obtained for word recognition using methods proposed in this work.
As it can be seen in the Figure \ref{fig:retrieved}, baseline word recognition system fails to recognise some of the minute detials.
For example, in Figure \ref{fig:retrieved}(a) and (d) the baseline word recognition system fails to predict the `.' present in the words.
Whereas, in Figure \ref{fig:retrieved} (b) and (c) the baseline recognition system fails to identify a couple of \textit{matras}.
Using the methods proposed in this work, we are able to capture the minute details missed by the baseline work recognition system and contribute towards improving the word recognition system.

All the methods proposed in this work are analysed on the Hindi language.
Though our experimental validation is limited to printed Hindi 
books, we believe, the methods are generic, and 
independent to underlying recognition and embedding networks.
Word retrieval systems using word spotting are known to have a higher recall, whereas, text recognition-based systems provide higher 
precision~\cite{main}. Exploring the aforementioned fact, in this work, we propose various ways of fusing text recognition and word spotting based systems for improving word recognition and retrieval.


\subsection{Related works}
We build upon the recent work from text recognition and word spotting.
We propose a new set of methods which improves these existing methods by using their complementary properties.
In this section we explore the work done in (a) text recognition and (b) word spotting.

Modern text recognition solutions are  typically modelled as a Seq2Seq problem using Recurrent Neural Networks ({\sc rnn}s). 
In such a setting, convolutional layers are used for extracting features, leading to Convolutional Recurrent Neural Network ({\sc crnn}) based solutions. 
Transcriptions for the same are generated using different forms of recurrent networks.
For example, Garain et al. \cite{Garain} use {\sc blstm}s along with {\sc ctc} loss \cite{ctc} and Adak et al. \cite{Adak} use {\sc cnn}s as features extractors coupled with {\sc rnn} for sequence classification.
Similarly, Sun et al. \cite{Sun} use a convolutional network as the feature extractor and multi-directional ({\sc md}ir) {\sc lstm}s as the recurrent units. 
Pham et al. \cite{Pham} also use the convolutional network as feature extractor whereas they use multi-dimensional long short-term memory ({\sc mdlstm}) as the recurrent units. 
Chen et al. \cite{Chen} use a variation of the {\sc lstm} unit, which they call Sep{\sc mdlstm}.
All these methods heavily rely on recognition-based approach for word recognition and do not explore recognition-free approaches for improving word recognition and retrieval.
Whereas, we use a {\sc cnn-rnn} hybrid architecture first proposed in \cite{Shi} for generating the textual transcriptions.
To add to it we use the End2End network \cite{e2e} for generating deep embeddings of the transcriptions to further improve the word recognition using methods defined in Section 3.


In word spotting the central idea is to compute the holistic representation for the word image.
Initial methods can be traced back to \cite{Rath} where Rath et al. uses profile features to represent the word images and then compared them using a distance metric. 
Many deep learning approaches like \cite{Jaderberg_2,Jaderberg_3} have been proposed in the domain of scene text recognition for improved text spotting. 
Poznanski et al. \cite{Poznanski} adopted {\sc vggn}et \cite{VGG} by using multiple fully connected layers, for recognising attributes of Pyramidal Histogram of Characters ({\sc phoc}). 
Different {\sc cnn} architectures \cite{Kris,Sudholt,Sudholt_1,Wilkinson} were proposed which uses {\sc phoc} defined embedding spaces for embedding features. 
On the other hand, Sudholt et al. \cite{Sudholt} suggested an architecture which embeds image features to {\sc phoc} attributes by using sigmoid activation in the last layer. 
It uses the final layer to get a holistic representation of the images for word spotting and is referred to as {\sc phocn}et.
Methods prior to deep learning use handcrafted features \cite{Balasubramanian,Meshesha}, and Bag of Visual Words approaches \cite{Shekhar} for comparing the query image with all other images in the database.
Various attempts of retrieving Indian texts using word spotting methods have been illustrated in \cite{Bhardwaj,Chaudhury,Meshesha,Shekhar}.
All these methods greatly explore the recognition-free approach for word recognition. 
They do not study the fusion of recognition-free approach with recognition-based approaches to use the complementary information from both the systems.

In this paper, we explore various methods using which we are able to fuse both recognition-based and recognition-free approaches.
We convert the text transcriptions generated by recognition-based methods to deep embeddings using recognition-free approach and improve word recognition and retrieval.
Our major contributions are:
\begin{enumerate}
    \item We propose techniques for improving word recognition by incorporating multiple hypotheses and corresponding deep embeddings.
    Using this technique, we report an average improvement in the word accuracy by $1.4\%$.
    \item 
    Similarly for improving word retrieval we propose techniques using both recognition-based and recognition-free approaches.
    Using approaches introduced in this paper, we report an average improvement in the mAP score by $11.12\%$.
    We release an implementation at our webpage\footnote{\href{http://cvit.iiit.ac.in/research/projects/cvit-projects/fused-text-recogniser-and-deep-embeddings-improve-word-recognition-and-retrieval}{http://cvit.iiit.ac.in/research/projects/cvit-projects/fused-text-recogniser-and-deep-embeddings-improve-word-recognition-and-retrieval}}.
\end{enumerate}{}



\section{Baseline Methods}
We first explain the two baseline methods that we are using in this work for our task.

\subsection{Text Recognition}
\label{sec:ocr}

\begin{figure}[h!]
    \centering
    \includegraphics[width=\textwidth]{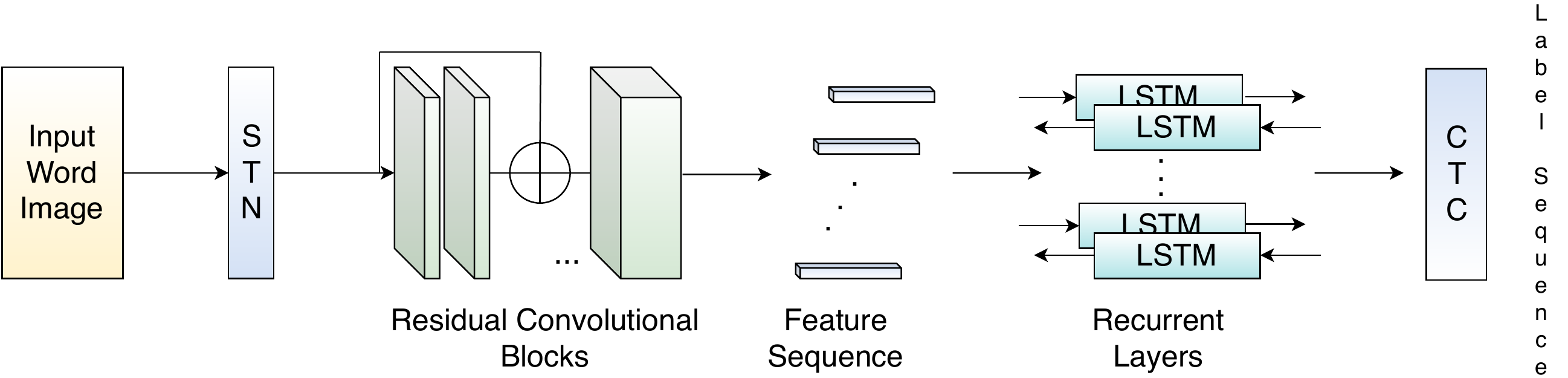}
    \caption{{\sc crnn} architecture takes in a word image passes it through the Spatial Transform ({\sc stn}) layer (for correcting the affine transformation applicable to the word images), followed by residual convolutional blocks which learns the feature maps. These feature maps are further given as an input to the {\sc blstm} layer.}
    \label{fig:CRNNArc}
\end{figure}{}

The problem of text recognition involves converting the content of an image to textual transcriptions. 
For this purpose, we use {\sc cnn-rnn} hybrid architecture which was first proposed by \cite{Shi}. 
Figure \ref{fig:CRNNArc} shows the {\sc crnn} with the {\sc stn} layer proposed in \cite{crnn_modified}.
Here, from the last convolutional layer we obtain a feature map $F_l \in \rm I\!R^{\alpha\times\beta\times\gamma}$ which is passed as an input to the {\sc blstm} layers as a sequence of $\gamma$ feature vectors, each represented as $F_{l+1} \in \rm I\!R^{\alpha\times\beta}$.
Here $l$ is the layer {\sc id}.
{\sc ctc} loss \cite{ctc} is used to train the network with best path decoding.

\subsection{Word Spotting}

\begin{figure}[h!]
    \centering
    \includegraphics[scale=0.5]{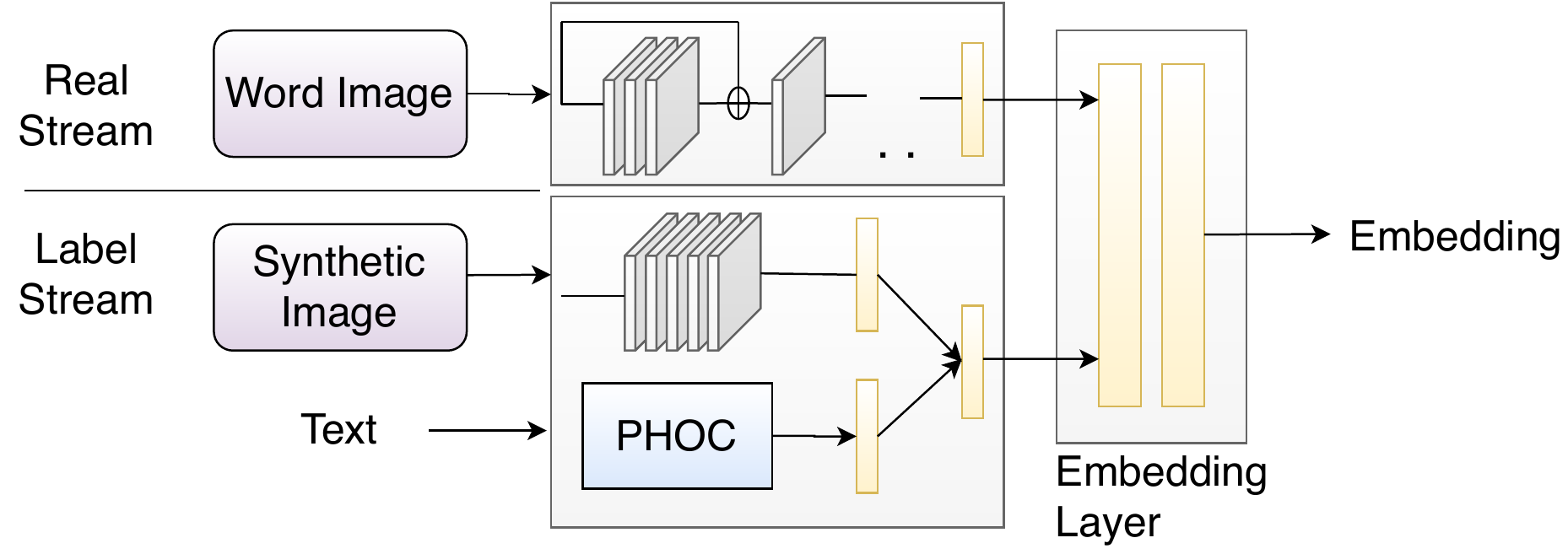}
    \caption{End2End\cite{e2e} network for learning both textual embedding using a multi-task loss function.}
    \label{fig:E2E}
\end{figure}{}

We use the End2End network proposed in \cite{e2e} to learn the textual and word image embeddings. Figure \ref{fig:E2E} shows this architecture. Feature extraction and embedding are the major components of the network. The network consists of two input streams - Real Stream and Label Stream as shown in Figure \ref{fig:E2E}. The real stream takes in real-word images as input and feeds it into a deep residual network which computes the features. The label stream gets divided into two different streams - the {\sc phoc} \cite{Almazan} feature extractor and a convolutional network. A synthetic image of the current label is given as an input to the convolutional network which in turn calculates its feature representation. This feature representation is in turn concatenated with the vectorial representation which is calculated using {\sc phoc}. Features generated from both the streams are fed to the label embedding layer which is responsible for projecting the embeddings in a common feature space where both the embeddings are in close proximity.


\section{Fusing Word Recognition and Word Spotting}
We propose a system that leverages the best traits of both the recognition-based and recognition-free approaches.
Given the document images, word images are cropped out using the word bounding box information available as part of the annotation.
The word images are fed into a pre-trained text recognition network as described in  the Section 2.1 to get the textual transcriptions. 
Similarly, the word images are also fed into the End2End network \cite{e2e} for getting word images' deep embeddings. 

\titlerunning{}
\begin{figure}[h!]
    \centering
    \includegraphics[scale=0.45]{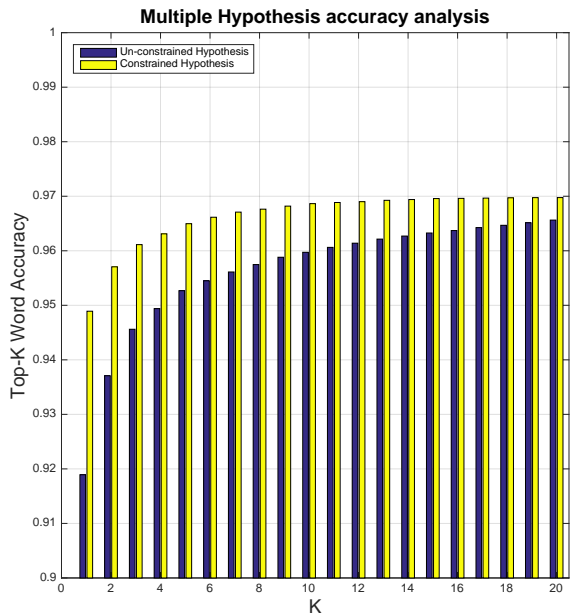}
    \caption{Top-K word accuracies of $K$ hypotheses generated using beam search decoding algorithm on {\sc crnn}'s output. Where $K$ is the number of hypotheses generated by the text recogniser.}
    \label{fig:topk}
\end{figure}



\subsection{Word Recognition using Multiple Hypotheses} 
To improve word recognition, we use the beam search decoding algorithm for generating $K$ hypotheses from the text recognition system.
Figure \ref{fig:topk} shows a graph of top-k word accuracy vs. $K$, where $K$ is the number of hypotheses generated by the text recogniser.
Bars in dark blue show the top-k accuracy. 
Here we consider the output to be correct if the correct word occurs once in $K$ outputs.
We can further improve the word accuracy by imposing a constraint with the help of a lexicon and removing hypotheses which are not part of the lexicon.
By doing this we reduce the noise in our $K$ hypotheses.
The lexicon, in our case, was taken from \cite{iitkgpLink}.
Yellow bars in Figure \ref{fig:topk} show the top-k accuracy after filtering predictions using the lexicon.
In this work we devise methods using which we can select the best hypothesis from these $K$ hypotheses.
We use deep embeddings generated by the End2End network. 
All the $K$ hypotheses are passed through the End2End network \cite{e2e} and their deep embeddings are denoted by $E_{m_j}\; \forall j \in\{1, ..., K\}$. 

\subsubsection{Notation}
Given a dataset of $n$ word images their embeddings are denoted by $E_{w_i} \; \forall i\in\{1, ..., n\}$. $E_{img}$ represents the embedding of the word image we want to recognise.
All the text inputs are passed through the label stream shown in the Figure \ref{fig:E2E}.
Input query text, text recogniser's noisy output and text recogniser's multiple hypotheses when converted to embeddings (using the label stream) are denoted by $E_{t}$, $E_{n_i} \; \forall i\in\{1, ..., n\}$, and $E_{m_j}\; \forall j\in\{1, ..., K\}$ respectively.

\titlerunning{}
\begin{figure}[h!]
    \centering
    \includegraphics[scale=0.36]{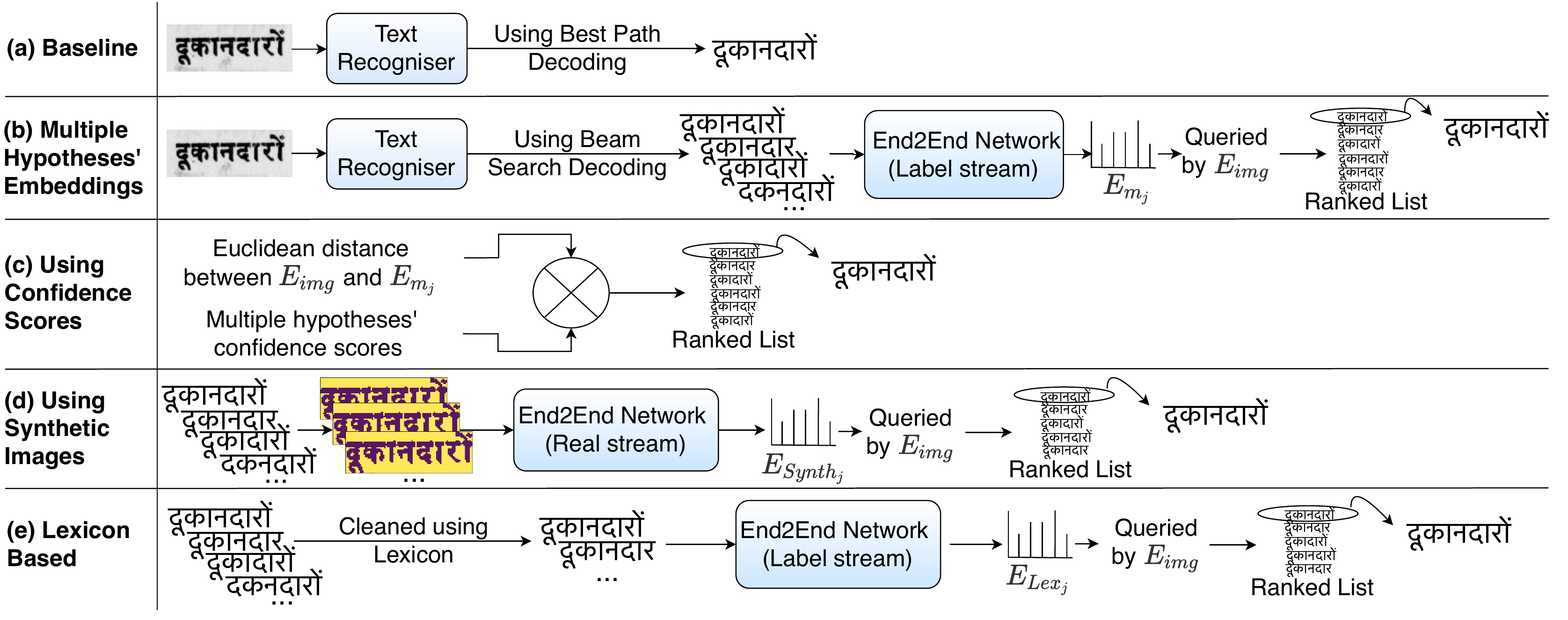}
    \caption{(a) A single output text is generated using the best path decoding algorithm, (b) multiple hypotheses are generated using the beam search decoding algorithm, which are passed through the End2End network. The embedding generated ($E_{m_j}$) is queried by $E_{img}$ to get a ranked list of predictions. Prediction at the first position is the new output. (c) Distance between $E_{img}$ and $E_{m_j}$ is summed with the confidence score to generate the ranked list. (d) Synthetic images are generated corresponding to the multiple hypotheses which are then converted to embedding ($E_{Synth_j}$) using the End2End network. $E_{Synth_j}$ is queried by $E_{img}$ to get a ranked list of predictions. (e) Here multiple hypotheses are limited using a lexicon, which is converted to embedding ($E_{Lex_j}$) using the End2End network. $E_{Lex_j}$ is queried by $E_{img}$ to get a ranked list.}
    \label{fig:Word_recognition_pipeline}
\end{figure}


\subsubsection{Baseline word recognition system} Figure \ref{fig:Word_recognition_pipeline}(a) shows the baseline method that generates a single output. 
The baseline results for constrained and unconstrained approaches can be seen at $K=1$ in Figure \ref{fig:topk}. 
It can be concluded from Figure \ref{fig:topk} that, correct word is more likely to be present as $K$ increases.
In this section we present methods which allows us to improve the word recogniser's output by selecting the correct word out of $K$ predictions.

\subsubsection{Using multiple hypotheses' embeddings} As it can be seen in Figure \ref{fig:Word_recognition_pipeline}(b), an input image is given to the text recogniser.
The text recogniser uses beam search decoding algorithm for generating multiple hypotheses for the input image.
These hypotheses are then passed through the label stream of the End2End network to generate $E_{m_j}$.
$E_{m_j}$ is queried by $E_{img}$ to get a ranked list.
The string at the top-1 position of the ranked list 
is considered to be the text recogniser's new output.

\subsubsection{Using confidence scores} A confidence score is associated with $E_{m_j}$ generated by the beam search decoding algorithm. 
This method uses that confidence score to select a better prediction.
As seen in Figure \ref{fig:Word_recognition_pipeline}(c) the Euclidean distance between $E_{img}$ and $E_{m_j}$ is summed with the confidence score. 
This new summed score is used to re-rank the ranked list obtained by querying $E_{img}$ on $E_{m_j}$. 
The string at the top-1 position of the re-ranked list is considered to be the text recogniser's new output.

\subsubsection{Using synthetic images} In this method we bring $E_{m_j}$ closer to $E_{img}$ by converting multiple hypotheses to synthetic images\footnote{Generated using \href{https://pango.gnome.org}{https://pango.gnome.org}.}. 
Here, we exploit the fact that in the same subspace, the synthetic image's embeddings will lie closer to $E_{img}$ as compared to $E_{m_j}$ (text embeddings).
As shown in Figure \ref{fig:Word_recognition_pipeline}(d), we generate synthetic images for the multiple hypotheses provided by the text recognition system.
These synthetic images are then passed through the End2End network's real stream to get the embeddings denoted by $E_{synth_j}\; \forall j\in\{1, ..., K\}$. 
$E_{synth_j}$ is queried using $E_{img}$ to get a ranked list which contains the final text recognition hypothesis at top-1 position. 
This method performs better as compared to the one using $E_{m_j}$ because image embeddings will be closer to the input word-image embedding as they are in the same subspace.

\subsubsection{Lexicon based recognition} Figure \ref{fig:topk} shows that using lexicon-based constrained hypotheses, we can get much better word accuracy as compared to the hypotheses generated in an unconstrained fashion. 
In this method, we exploit this fact and limit the multiple hypotheses generated from the text recognition system. 
As shown in Figure \ref{fig:Word_recognition_pipeline}(e), we limit the multiple hypotheses after they are generated.
These filtered hypotheses are then passed through the label stream of the End2End network.
The embeddings generated are denoted by $E_{Lex_j}\; \forall j\in\{1, ..., K\}$.
$E_{m_j}$ is now replaced by $E_{{Lex}_j}$.
This method outperforms the previous methods as it decreases the noise in the hypotheses by constraining the number of words to choose from.

\subsection{Word Retrieval using Fusion and Re-Ranking}

For creating a retrieval system, we get the deep embeddings for the input query text ($E_{t}$) and also for the word images ($E_{w_i}$) from the documents on which we wish to query the input text. 
To use the best traits of the text recognition system, we also convert the text generated by the text recognition system to deep embeddings ($E_{n_i}$). 
In this section, we propose various techniques for improving word retrieval.

\titlerunning{}
\begin{figure}[h!]
    \centering
    \includegraphics[width=\textwidth]{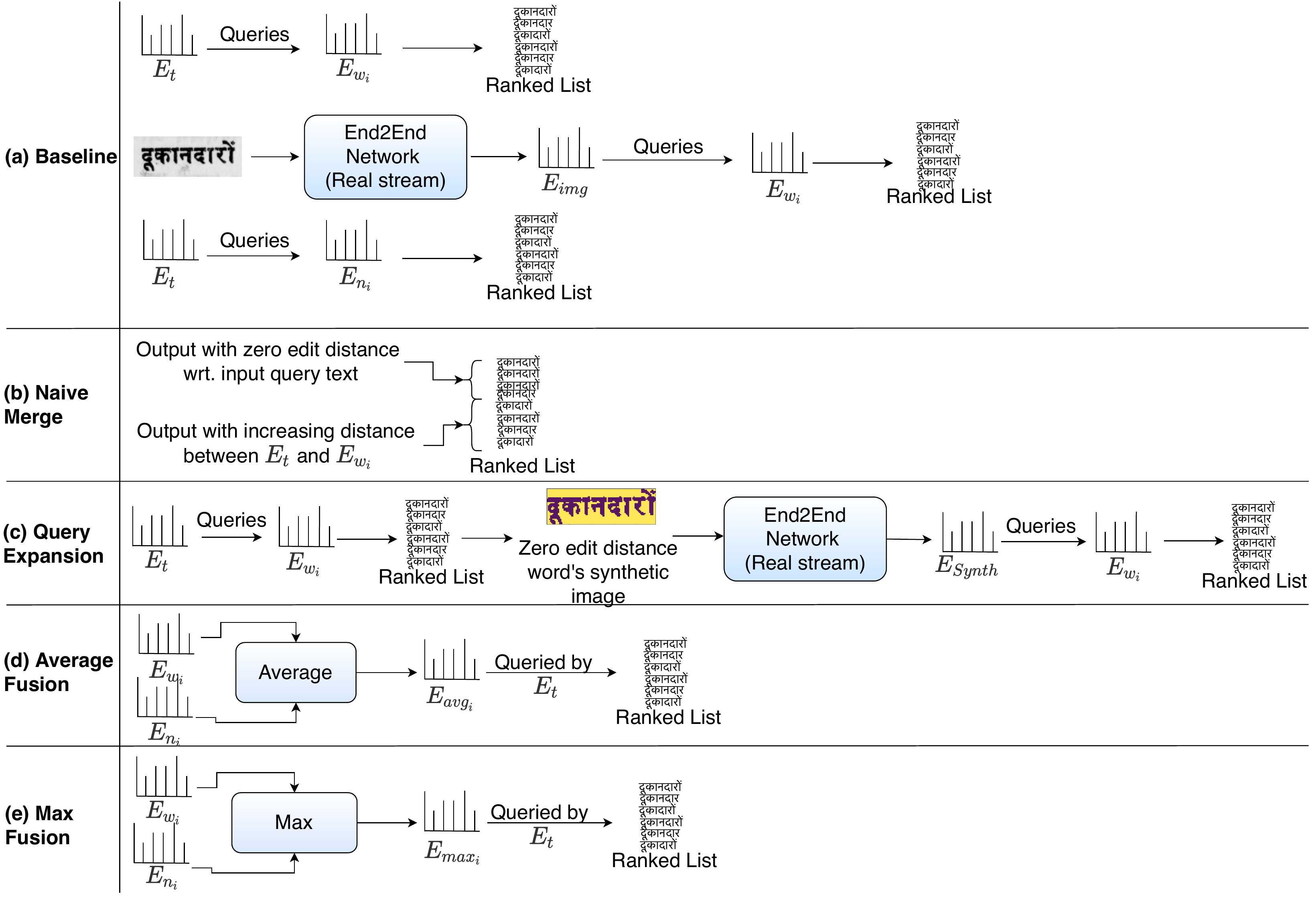}
    \caption{(a) Baseline for word retrieval queries $E_{t}$ on $E_{w_i}$ and $E_{n_i}$. (b) Naive Merge attempts to exploit the best traits of text recognition and word spotting. (c) In Query Expansion, we merge the “query by string" and “query by example" setting. (d),(e) Average and Max Fusion methods fuse $E_{w_i}$ and $E_{n_i}$ by performing average and max operation respectively.}
    \label{fig:Word_retreival_pipeline_1}
\end{figure}

\subsubsection{Baseline Word Retrieval System} There are three major ways to perform baseline word retrieval experiment. 
The first method focuses on demonstrating the word retrieval capabilities of embeddings in a “query by string" setting. 
Shown in Figure \ref{fig:Word_retreival_pipeline_1}(a), $E_{t}$ is queried on $E_{w_i}$ to get a ranked list. 
The second method focuses on demonstrating the word retrieval capabilities of embeddings in a “query by example" setting.
Here input is a word image which is converted to embedding $E_{img}$ using the real stream of the End2End network.
$E_{img}$ is used to query $E_{w_i}$ to get a ranked list.
The third method focuses on demonstrating the word retrieval when done on $E_{n_i}$ in a “query by string" setting. 
Here, $E_{t}$ is queried on $E_{n_i}$ to get a ranked list. 
This method is a measure of how well the text recogniser is performing without the help of $E_{w_i}$. 
The performance of this method is directly proportional to the performance of the text recogniser.

Another way in which we can measure the performance of text recogniser's noisy output is by using the edit distance.
Edit distance measures the number of operations needed to transform one string into another. We calculate the edit distance between input query text and text recogniser's noisy output. The ranked list is created in an increasing order of edit distance. This experiment gauges the contribution of the text recogniser in the word retrieval process.

\subsubsection{Naive Merge} As shown in the Figure \ref{fig:Word_retreival_pipeline_1}(b), we get the initial ranked list by calculating the edit distance between input text and noisy text recogniser's output.
The remaining words are arranged in the order of increasing Euclidean distance between $E_{t}$ and $E_{w_i}$.
This method exploits the best traits of both the methods - text recognition and word spotting for creating a ranked list.\\[-4.5 ex]

\subsubsection{Query Expansion} 
In this method, initially we use the “query by string” setting and in the second stage “query by example” setting is used.
The “query by example” setting works much better than the “query by string” setting, as image embedding is used to query on $E_{w_i}$ and image embeddings lie closer in the subspace. 
As shown in Figure \ref{fig:Word_retreival_pipeline_1}(c), from the initial ranked list, we get a word with zero edit distance. 
A synthetic image corresponding to this word is generated.
It is then fed to the real stream of the End2End network.
We then get that word's synthetic image embedding ($E_{Synth}$).
It is then queried on $E_{w_i}$ to get a re-ranked list.
The only case in which this method can fail is, when the text recogniser generates a wrong word with respect to the input image and the generated word matches with the input query; in this particular case, we end up selecting an incorrect image embedding for re-ranking and get a wrong ranked list.

\subsubsection{Average Fusion} In this method, we merge $E_{w_i}$ and $E_{n_i}$. As shown in \ref{fig:Word_retreival_pipeline_1}(d), we perform average of corresponding $E_{w_i}$ and $E_{n_i}$, which is called the averaged embedding ($E_{avg_i}$). The $E_{t}$ is then queried on $E_{avg_i}$ to get a ranked list.

\subsubsection{Max Fusion} In this method, we merge $E_{w_i}$ and $E_{n_i}$. As shown in \ref{fig:Word_retreival_pipeline_1}(e), we the output having a maximum value between $E_{w_i}$ and $E_{n_i}$, which is called the max Embedding ($E_{max_i}$). The $E_{t}$ is then queried on $E_{max_i}$ to get a ranked list.

\section{Experiments}
In this section, we discuss the dataset details and results on the experiments described in Section 3.

\subsection{Dataset and evaluation metrics}

\begin{table}[h!]
    \caption{Dataset Details}
    \centering
    \begin{tabular}{|c|c|c|c|c|}
        \hline
         Dataset name & Annotated & \#Pages & \#Words & Usage\\
         \hline
         $Dataset1$ & Yes & 1389 & 396087 & Training\\
         \hline
         $Dataset2.1$ & Yes & 402 & 105475 & Word Recognition\\
         \hline
         $Dataset2.2$ & Yes & 500 & 120000 & Word Retrieval\\
         \hline
    \end{tabular}
    \label{tab:dataset}
\end{table}{}

We use two type of data collections for implementing and evaluating various strategies for word retrieval and recognition. 
In the first collection (Dataset1), the books were scanned and annotated internally.
In the second collection (Dateset2) the books were randomly sampled from the {\sc dli} \cite{dli} collection. 
Books in this collection range from different time periods.
They consists of variety of font sizes and are highly degraded.
Dataset2 is further divided into Dataset2.1 and Dataset2.2 for word recognition and word retrieval respectively.
The word retrieval experiments are performed using $17,337$ queries, which is a set of unique words from $500$ pages selected.
Dataset 2.1 and 2.2 have some pages common among them.
Table \ref{tab:dataset} summarises the datasets used in this work. 
To get models which generalize well and are unbiased towards any particular dataset, we train our End2End network and word recogniser on Dataset1 and test the models on Dataset2.
Both of these datasets contain annotated books.

We evaluate our word recognition system in terms of word accuracy which is $1 - WER$ (Word Error Rate), where $WER$ is defined as $\frac{S + D + I}{S + D + C}$. Here $S$ is the count of substitutions, $D$ is the count of deletions, $I$ is the count of insertions and $C$ is the count of correct words.
All our word retrieval methods are evaluated using the mAP score, which is defined as $mAP = \frac{\sum_{q=1}^{Q} AvgP(q)}{Q}$. Here $Q$ is the number of queries, $AvgP(q)$ is the average precision for each query.

\subsection{Word Recognition}
\begin{figure}[h!]
    \centering
    \includegraphics[scale=0.5]{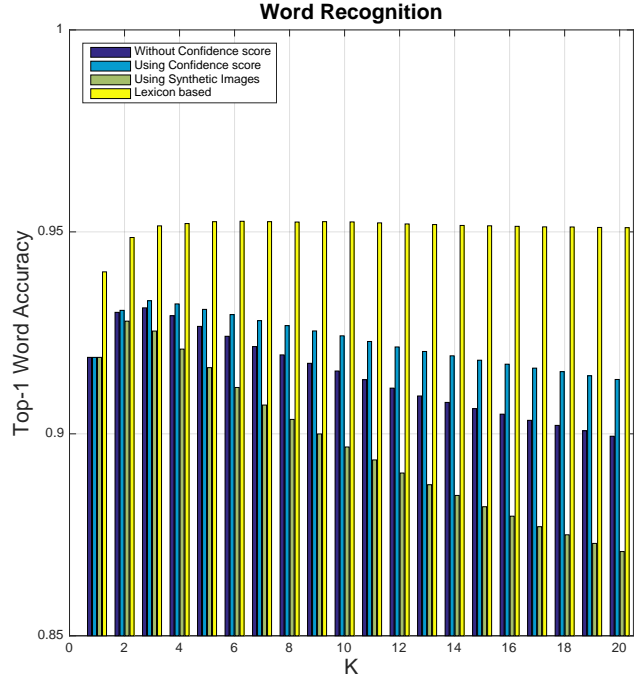}
    \caption{Word recognition results. Here K denotes the number of hypotheses generated by the text recogniser.}
    \label{fig:WordRecognition}
\end{figure}{}

This section shows the results obtained by applying the techniques discussed in Section 3.1 for word recognition on Dataset2.1.
There is no pre-processing done on these words and contain a high amount of noise, like presence of skewing, marks and cuts. 
Figure \ref{fig:WordRecognition} shows the results of word recognition experiments. 
The bars in dark blue colour show the variation in the word accuracies as $K$ is increased for the experiment in which we use the multiple hypotheses' embeddings as shown in Figure \ref{fig:Word_recognition_pipeline}(b).
It is evident as the $K$ is increased, at first, the word accuracy increases and then, starts to decrease; the reason is as $K$ increases the noise starts to increase and the algorithm tends to choose incorrect predictions. 
In this method for $K=3$, we report the highest word accuracy that is $93.12\%$, which is $1.2\%$ more than accuracy at $K=1$.

The bars in light blue show the results for the method where we incorporate the confidence information.
It shows a higher gain in the accuracy score as $K$ is increased. 
At $K=3$, we report the highest word accuracy, which is $1.4\%$ more than the accuracy at $K=1$.
Green bars in Figure \ref{fig:WordRecognition} show the result for experiment using synthetic images.
At $K=2$ we report the highest word accuracy which is $0.65\%$ more than the accuracy at $K=1$.
The lexicon-based method shows the best result of all with the highest accuracy reaching to $95.26\%$, which is $1.25\%$ more as compared to $K=1$ (shown using yellow bars in Figure \ref{fig:WordRecognition}). 
And it can be observed that the results are more consistent in this case, and don't drop as they do for the other methods; the reason being there is a very little amount of noise in the top predictions.

\subsection{Word Retrieval}
This section shows the results obtained by applying the techniques discussed in Section 3.2 for word retrieval on $Dataset2.2$.
In table \ref{table:wordretrieval}, rows 1-4 show the baseline results for word retrieval. 
The mAP of the ranked list generated by calculating edit distance on text recogniser's noisy output is $82.15\%$, whereas mAP for query by string (QbS) method on text recogniser's noisy word embeddings is $90.18\%$. 
This proves that converting text recogniser's noisy output to deep embeddings ($E_{n_i}$) helps in 
capturing the
most useful information. 
The mAP of the ranked list generated by querying $E_{t}$ on $E_{w_i}$ is 92.80\%; this shows that $E_{n_i}$ is noisier as compared to $E_{w_i}$. 
These results lead us to the conclusion that we can use the complementary information from the text recogniser's noisy text and word images for improving word retrieval.
Highest mAP is observed in the query by example (QbE) setting which is $96.52\%$. 
As we have used image input for querying we get the highest mAP score in this case, whereas in other cases we are using string as the input modality.

\begin{table}[ht]
    \caption{This table summarises the results for all the word retrieval experiments. It can be observed that methods using “query by example" setting work best in the baseline as well as the re-ranking case. In the case of fusion, average fusion proves to be the one showing the most improvement.}
    \centering
    \begin{tabular}{| p{1.65cm} | p{1.5cm} | p{6.5cm} | p{1.3cm}|}
    \hline
    Experiment Type & Input Modality & Experiment & mAP\\
    \hline
    \multirow{4}{*}{Baseline} & String & Edit Distance on Text Recogniser's Outputs & 82.15 \\
    \cline{2-4}
    & String & QbS on Text Recogniser's Embeddings & 90.18\\
    \cline{2-4}
    & \textbf{String} & \textbf{QbS Word Image Embeddings} & \textbf{92.80} \\
    \cline{2-4}
    & \textbf{Image} & \textbf{QbE Word Image Embeddings} & \textbf{96.52} \\
    \hline
    \multirow{2}{*}{Re-ranking} & String & Naive merge & 92.10 \\
    \cline{2-4}
    & \textbf{String} & \textbf{Query Expansion} &  \textbf{93.18} \\
    \hline
    \multirow{2}{*}{Fusion} & \textbf{String} & \textbf{Average Fusion} & \textbf{93.07}\\
    \cline{2-4}
    & String & Max. Fusion & 92.79 \\
    \hline
    \end{tabular}
    \label{table:wordretrieval}
\end{table}

Row 5-6 in table \ref{table:wordretrieval} shows the results obtained by re-ranking the ranked lists. 
The mAP of the ranked list obtained by naive merge is $92.10\%$. 
Naive merge shows an improvement over baselines by using the best output of the text recogniser for the initial ranked list.
And arranging the remaining word images in increasing order of Euclidean distance between $E_t$ and $E_{w_i}$.
The mAP of the ranked list obtained by query expansion is $93.18\%$. 
Query expansion shows improvement over all the previous methods as here we get a second-ranked list by using the text recogniser's best output's word image.
It is intuitive that, if we use a word-image embedding for querying $E_{w_i}$ the mAP will improve.


Rows 7-8 in table \ref{table:wordretrieval} shows the results obtained by fusing $E_{n_i}$ and $E_{w_i}$. 
The ranked list obtained by $E_{avg_i}$ gives us an mAP score of $93.07\%$, and the one obtained by $E_{max_i}$ is $92.79\%$. 
The fusion methods show an improvement over the baselines in QbS as they contain information of both $E_{w_i}$ and $E_{n_i}$.

\subsection{Failure cases}
\begin{figure}[h!]
    \centering
    \includegraphics[scale=0.5]{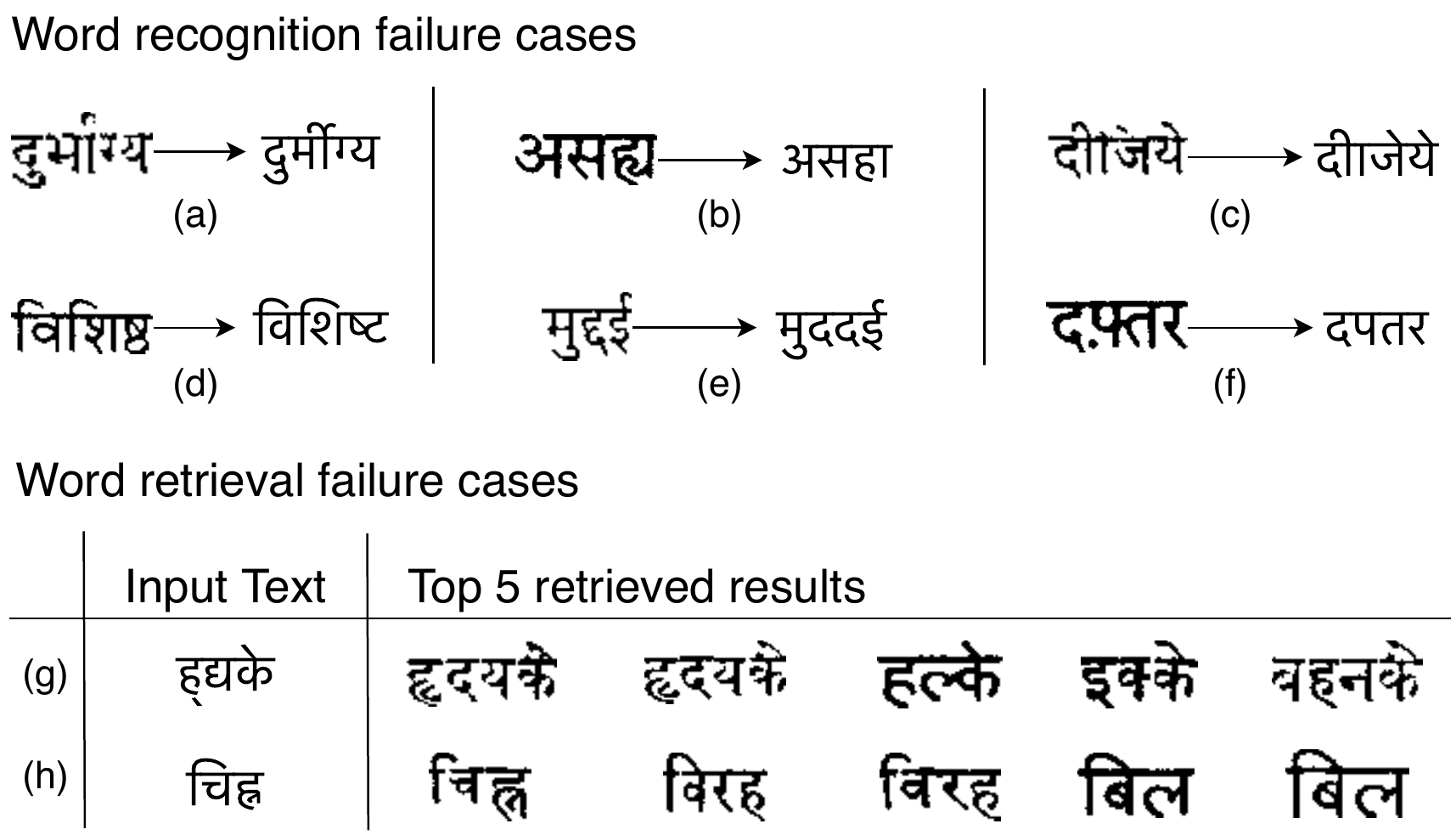}
    \caption{Failure cases}
    \label{fig:failure_cases}
\end{figure}

Figure \ref{fig:failure_cases} shows instances where both word recognition and retrieval fail to achieve the desired results.

{}

In case (a) and (c) the image is degraded due to which we are not able to capture the image information well and end up generating the wrong prediction. In case (b), (d), (e), and (f) the characters resemble closely to other characters which leads to a wrong output. To add to this, in the case of (b), (d), and (e) the characters are rare which increases the confusion. Cases (g) and (h) show the example where word retrieval was not able to perform well. For case (g) and (h), the input query has a rare character which resembles other characters, due to which the words are incorrectly retrieved. Though, for (h), using query expansion technique, we were successful in retrieving one out of two instances of the query.

\section{Conclusion}
To summarise, we  improve the word retrieval process by using the deep embeddings generated by the End2End network~\cite{e2e}. 
We have shown that by using the complementary information of text recogniser and word spotting methods, we can create word recognition and retrieval system capable of performing better than both of the individual systems.
We plan to explore various other fusion techniques apart from average and max fusion for improving word retrieval. 

\bibliographystyle{splncs04}
\bibliography{ref}
\end{document}